\documentclass{article}

\usepackage[preprint, nonatbib]{neurips_data_2022}
\usepackage[utf8]{inputenc} % allow utf-8 input
\usepackage[T1]{fontenc}    % use 8-bit T1 fonts
\usepackage[backref=section]{hyperref}       % hyperlinks
\usepackage{url}            % simple URL typesetting
\usepackage{booktabs}       % professional-quality tables
\usepackage{amsfonts}       % blackboard math symbols
\usepackage{nicefrac}       % compact symbols for 1/2, etc.
\usepackage{microtype}      % microtypography
\usepackage[dvipsnames]{xcolor}  % colors
\usepackage{subfiles}
\usepackage{graphicx}
\usepackage{multirow}
\usepackage{float}          % table positioning
\usepackage{wrapfig}
\usepackage{makecell}

\usepackage{amsmath}
\usepackage{amssymb}
\usepackage{xspace}

\definecolor{cbblue}{RGB}{0, 114, 178}
\definecolor{cbgreen}{RGB}{0, 158, 115}
\definecolor{cbred}{RGB}{213, 94, 0}
\definecolor{cbbrown}{RGB}{124, 65, 28}

\newcommand{\blue}[1]{{\color{cbblue}#1}}
\newcommand{\red}[1]{{\color{cbred}#1}}
\newcommand{\green}[1]{{\color{cbgreen}#1}}
\newcommand{\brown}[1]{{\color{cbbrown}#1}}

\newcommand{\ma}[1]{{#1}}

\newcommand{\maa}[1]{{#1}}

\def\IVMw{IVM$_{wthn}$\xspace}
\def\IVMb{IVM$_{btwn}$\xspace}
\def\AB{A\&B\xspace} % shortcut for Ackerman and Ben-David
\def\CH{$CH$\xspace}
\def\CHb{$CH_{btwn}$\xspace}
\def\CHv{$CH_v$\xspace}

\newcommand{\rev}[1]{#1}

\usepackage{soul}

\title{Sanity Check for External Clustering Validation Benchmarks using Internal Validation Measures}

\author{
    Hyeon Jeon$^1$ $\quad$ Micha\"el Aupetit$^2$ $\quad$ DongHwa Shin$^1$ $\quad$ Aeri Cho$^1$ $\quad$  \\ \textbf{Seokhyeon Park}$^1$ $\quad$  \textbf{Jinwook Seo}$^1$ \\
    $^1$Seoul National University, Seoul, Korea \\
    $^2$Qatar Computing Research Institute, Hamad Bin Khalifa University, Doha, Qatar\\
    \texttt{\{hj, dhshin, archo, shpark\}@hcil.snu.ac.kr, }\\ \texttt{maupetit@hbku.edu.qa, jseo@snu.ac.kr}
}

\begin{document}

\maketitle

\begin{abstract}
We address the lack of reliability in benchmarking clustering techniques based on labeled datasets. A standard scheme in external clustering validation is to use class labels as ground truth clusters, based on the assumption that each class forms a single, clearly separated cluster. However, as such cluster-label matching (CLM) assumption often breaks, the lack of conducting a sanity check for the CLM of benchmark datasets casts doubt on the validity of external validations. Still, evaluating the degree of CLM is challenging. For example, internal clustering validation measures can be used to quantify CLM \ma{within the same dataset} to \ma{evaluate} its different clusterings \ma{but are not designed to compare clusterings of different datasets}. In this work, we propose a principled way to generate between-dataset internal measures that enable the comparison of CLM across datasets. We first determine four axioms for between-dataset internal measures, complementing Ackerman and Ben-David's within-dataset axioms. We then propose processes to generalize internal measures to fulfill these new axioms, and use them to extend the widely used Calinski-Harabasz index for between-dataset CLM evaluation. Through quantitative experiments, we (1) verify the validity and necessity of the generalization processes and (2) show that the proposed between-dataset Calinski-Harabasz index accurately evaluates CLM across datasets. Finally, we demonstrate the importance of \maa{evaluating} CLM \ma{of benchmark datasets before} conducting external validation. 
\end{abstract}

\section{Introduction}

\label{sec:intro}

Cluster analysis \cite{jain88prenticehall} is an essential exploratory task for data scientists and practitioners in various application domains \cite{lyi15bioinformatics, schaeffer07gc, caron18eccv}. It relies on clustering techniques, that is, unsupervised machine learning algorithms that partition the data into subsets called groups or clusters, while maximizing between-cluster separation and within-cluster compactness based on some distance function \cite{liu10idcm}.

Clustering validation measures \cite{farber10multiclust} or quality measures \cite{bendavid08nips} have been proposed to evaluate clustering results. They can be \textit{internal} or \textit{external} \cite{rendon11ac, liu10idcm, jain88prenticehall}. 
Internal validation measures (IVM) give high scores to partitions in which data points with high or low similarities to each other are assigned to the same or different clusters, respectively. In contrast, External validation measures (EVM) quantify how well a clustering matches a ground truth partition. \maa{Taking the classes of labeled data as ground truth is a widely used approach to rank clustering techniques on benchmark datasets \cite{farber10multiclust}.}%We are interested in the case where the ground truth is given as the classes of a labeled dataset, which is .

% in the widely used approach which gives the ground truth partition as the classes of a labeled dataset \cite{farber10multiclust}.

\begin{figure}[t]
    \centering
    \includegraphics[width=0.88\textwidth]{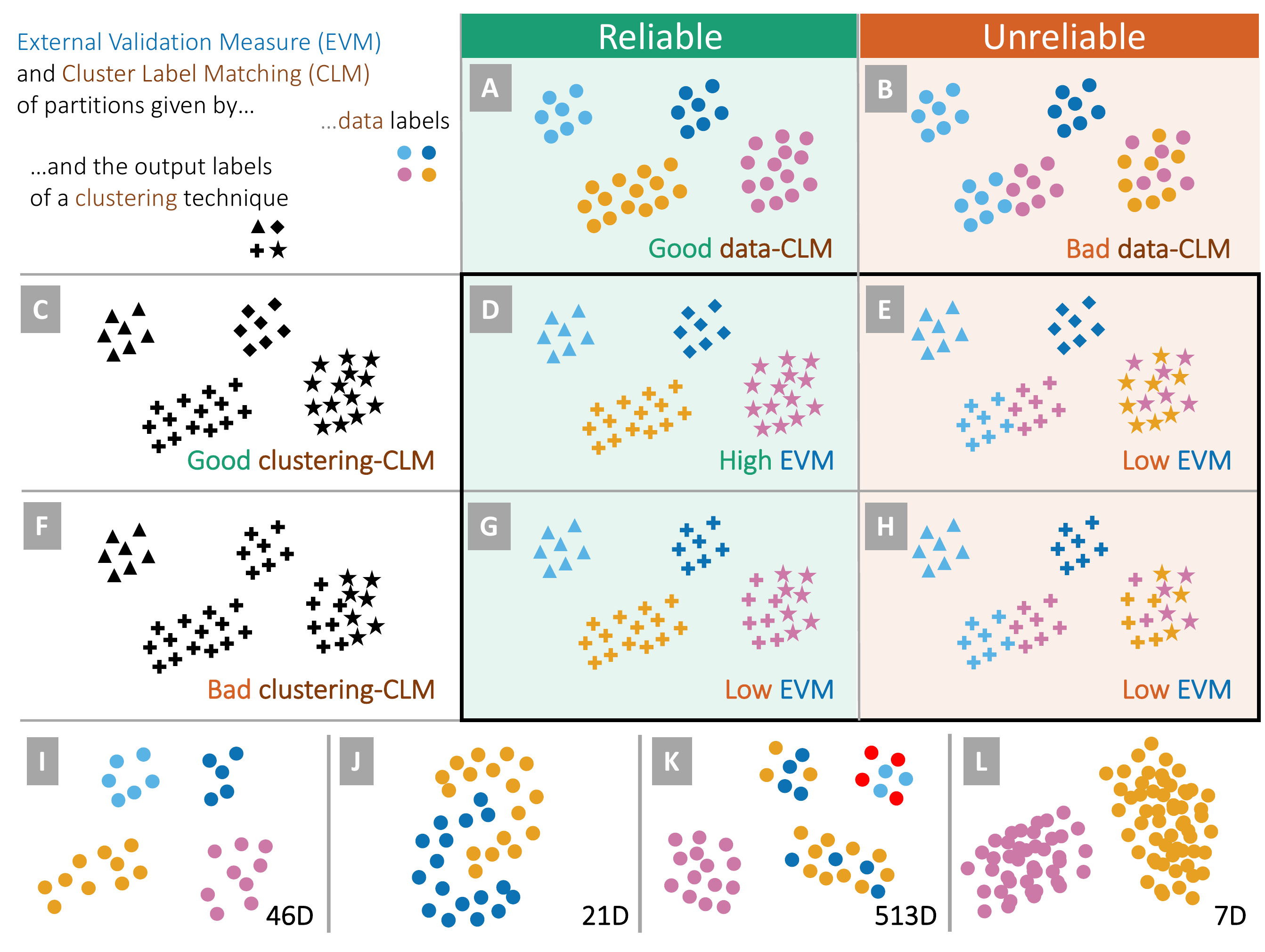}\\
    % \vspace{-0.5mm}
    \caption{An External Validation Measure (\blue{EVM}) evaluates the matching between two partitions (markers' shape and fill) of the same data (D, E, G, H). The Cluster-Label Matching (\brown{CLM}) is \green{good} (A, C) if the partition formed by the \textit{labels} assigned to the data (shape or fill) matches well the \textit{clusters} formed by the data point distribution (encoded by position). The labels can be provided with the \textit{data} (fill) (A, B) or as the output partition of the \textit{clustering} technique to be evaluated (shape)(C, F). If \brown{\textit{data-CLM}} is \green{good} (A), the \blue{EVM} between the data labels (A) and clustering labels (C, F) gives a \green{\textit{reliable}} evaluation (D, G) of the \brown{\textit{clustering-CLM}}: \green{high}/\red{low} \blue{EVM} (D/G) match with \green{good}/\red{bad} \brown{clustering-CLM} (C/F). But if \brown{data-CLM} is \red{bad} (B), the EVM is always \red{low} (E, H) and \red{\textit{unreliable}} to evaluate if the \brown{clustering-CLM} is good (C) or bad (F). It is highly unlikely to get a high EVM (as in D) if the \brown{data-CLM} is bad (B) as it would mean the clustering technique found by chance the same \textit{bad partition} as the one given by the data labels (fill (B) and shape (F) would match perfectly in H, not shown).
    \rev{Our main goal is to evaluate and compare \brown{data-CLM} of \textit{several datasets} (I,J,K,L) with different characteristics (dimension, size, data and class distributions)  %use only the best ones (I,L) 
    }\maa{to inform clustering evaluation with EVMs. }\rev{ But how can we get CLM scores comparable between datasets?}}
    \label{fig:clm}
    % \vspace{-4.1mm}
\end{figure}

\autoref{fig:clm} illustrates the main issue we propose to address in this work. Using class labels as ground truth in EVM relies on the Cluster-Label Matching (CLM) assumption that the dataset has a good matching between clusters and class labels \cite{aupetit14beliv} (\autoref{fig:clm}A). %However, this assumption often fails as classes might overlap with each other in one cluster, and points within one class may not form a compact cluster, \ma{\textit{i.e.} the dataset has a bad CLM} (\autoref{fig:clm}B) \ma{which makes EVM unreliable to rank clustering techniques based on such data (\autoref{sec:selection}).}
\maa{In the worst case, the CLM of the data can be bad with data ranging from having labels randomly assigned to or split between easy-to-detect clusters (\autoref{fig:clm}B), to having labels being well assigned to hard-to-detect clusters (\autoref{fig:clm}J). Then, a low EVM score can be due to a bad clustering of an otherwise good-CLM dataset (\autoref{fig:clm}G) (the clustering technique has low capacity to detect complex clusters) or to a bad-CLM dataset (\autoref{fig:clm}BJ) (the clustering technique may as well have a high capacity (\autoref{fig:clm}CE) or a low one (\autoref{fig:clm}FH), we cannot tell; EVM is unreliable when CLM is bad). Thus, it is crucial to evaluate CLM to measure the intrinsic quality of the ground truth dataset in order to inform and weigh the results of the EVM accordingly (\autoref{sec:selection}).} \rev{Still, the results of EVM over benchmark datasets are often given without considering their CLM \cite{rehioui16pcs, khan21icecit, monath17nips, monath19kdd} casting doubts on the rankings obtained}. \maa{Here, we aim to evaluate the CLM of labeled datasets}.

Yet, evaluating and comparing the CLM of benchmark datasets is challenging. We can use the \ma{average EVM scores of several} clustering techniques or the accuracy of classifiers in distinguishing classes \cite{rodriques18kbs, abul03smc} as a proxy for CLM, but such approaches are not based on principled axioms and are also very time-consuming. \ma{In contrast, IVMs are easy to compute, can derive from axioms~\cite{bendavid08nips} and} \ma{could be used as a proxy for CLM, as class partitions forming good clusters would get a higher score. 
However, they are designed to compare different clustering partitions of a \textit{single} dataset (\autoref{fig:clm} C \textit{vs} F), rather than the class partition of \textit{different} datasets (\autoref{fig:clm}IJKL). As datasets can differ by their size, dimension, data and class distributions, IVM scores of two labeled datasets \maa{cannot be reliably compared, making IVM improper measures of CLM across datasets; such claim is also confirmed by our experiments} (\autoref{sec:bdsa}; \autoref{tab:dataset_entire}). 
Thus, we lack a proper measure to compare CLM across datasets.} 

% compare different partitions of a single dataset; however, \rev{they are less accurate in evaluating and comparing \textit{across} datasets, as shown in our experiment (\autoref{sec:bdsa}; \autoref{tab:dataset_entire})}

% cannot evaluate and compare CLM \textit{across} datasets, \rev{}

In this research, we \ma{propose a set of new axioms from which we derive} a \textit{between-dataset} internal validation measure (IVM$_{btwn}$) as a grounded way to assess and compare CLM of different datasets.
\rev{An IVM$_{btwn}$ takes a single labeled dataset as input and returns a score evaluating its level of CLM. This score is designed to be comparable across datasets.}
%The measure enhances within-dataset internal to enable the comparison of CCM across datasets. 
Our contribution is four-fold: 

\textbf{Axioms } We propose four \emph{between-dataset axioms} that IVM$_{btwn}$  should satisfy for the fair comparison of CLM, complementing Ackerman and Ben-David's \emph{within-dataset} axioms \cite{bendavid08nips} satisfied by standard IVM (i.e., within-dataset IVM; \IVMw) (\autoref{sec:axiom}). These additional axioms require the IVM$_{btwn}$ to be invariant to the number of data points, classes, and dimensions, and to share a common range.

\textbf{Generalization process and new between-dataset Calinski Harabasz index} From these axioms, we propose technical tricks for generalizing an \IVMw into an \IVMb. We use them to generalize the Calinski-Harabasz index  ($CH$) \cite{calinski74cis} into  a \emph{between-dataset} $CH$ index  ($CH_{btwn}$) (\autoref{sec:ivmbtw}). 

% We demonstrate that both $CH$ and $CH_{btwn}$ satisfy all axioms \mac{Check details for this last claim}.

\textbf{Quantitative evaluations} Through an ablation study, we verify the validity and necessity of our generalization process (\autoref{sec:validity}). We also show that $CH_{btwn}$ \ma{ranking 96 real-world datasets significantly outperforms competitors in terms of rank-correlation with the ground truth CLM} \rev{approximated} \ma{based on nine clustering techniques, while being up to three orders of magnitude faster to compute than the} \rev{approximate} \ma{ground truth (Appendix D). These experiments demonstrate} \rev{the validity of our axiomatic approach and the effectiveness of $CH_{btwn}$ (\autoref{sec:bdsa}).}

% \textbf{Synthetic benchmark for validating IVM$_{btwn}$}. 
% We constructed benchmarks consist of 1,000 synthetic basis datasets each made of 10,000 samples from two Gaussian clusters with various covariance, means, and proportions, to experimentally validate whether IVM$_{btwn}$ satisfies the between-dataset axioms (\autoref{sec:validity}). We found that $CH_{btwn}$ passes the test, satisfying between-dataset axioms.

% \textbf{Quantitative Evaluations for between-dataset Calinski Harabasz Index.} Through additional quantitative experiments, we verified that $CH_{btwn}$ outperforms $CH$ and classifier-based EVM in terms of predicting the ground-truth CLM rank of 96 labeled datasets given by the best EVM score achieved among nine clustering techniques (\autoref{sec:bdsa}).

\textbf{Ranking real benchmark data for reliable EVM} Lastly, we explain the importance of evaluating CLM in advance of external validation by showing how not \maa{doing so} adversely affects the conclusions about the comparative performances of clustering techniques (\autoref{sec:selection}).

\section{\rev{Backgrounds and} Related Works} %Clustering Validation}

Many clustering techniques exist \rev{\cite{xu15ads}}, and ensemble approaches have been proposed to combine clustering results to compensate for the weaknesses of individual techniques \cite{vega11jprai}. Still, it is challenging to define what a \textit{good} clustering is.
For instance, stability is deemed an important criterion \cite{von10, ben06icclt}.  

%\subsection{External Validation}

EVM quantify how much the resulting clustering matches with a ground truth partition of the data. For example, Adjusted Mutual Information  \cite{strehl2002jmlr, vinh10jmlr} measures the agreement of two label assignments in terms of information gain corrected for chance effects. Other measures, such as Adjusted Rand Index \cite{santos09icann} or V-measure \cite{rosenberg07emnlp}, can be used instead. 

Classes of labeled datasets have been used extensively as ground truth for clustering EVM \cite{farber10multiclust}. %However, despite its potential risk of violating CLM, no principled procedure has yet been proposed to check the validity and reliability of using these benchmark datasets as ground truth, that is, to evaluate how much they satisfy the CLM assumption. Our research aims to fill this gap. 
\maa{However, despite its potential risk of violating CLM, no principled procedure has yet been proposed to evaluate the reliability of such a ground truth. Our research aims to fill this gap by proposing a measure of CLM. A similar endeavor has been engaged in the supervised learning community to quantify datasets' difficulty for classification tasks \cite{pmlr-v162-ethayarajh22a}.}

% Labeled datasets initially aimed to benchmark supervised classification techniques. For example, they have been used extensively as ground truth for clustering EVM \cite{farber10multiclust}, and to evaluate dimensionality reduction techniques \cite{aupetit14beliv, joia11tvcg}. However, no principled procedure has been proposed yet to check the validity and reliability of using these benchmark datasets as ground truth, \textit{i.e.} to evaluate how much they satisfy the CLM assumption. Our research aims to fill this gap. 

A natural idea would be to use classification scores as a proxy for CLM \cite{abul03smc, rodriques18kbs}. This approach is based on the assumption that the classes of a labeled dataset getting good classification scores will provide a reliable ground-truth for EVM. Still, a classifier is not designed to distinguish well between \ma{two "adjacent" classes forming a single cluster (\autoref{fig:clm}B light blue and purple bottom left cluster, good class separation but bad CLM) and two "separated" classes forming distant clusters (\autoref{fig:clm}A light blue and orange clusters, good CLM), nor it is designed to distinguish within-class structures like a class forming a single cluster (\autoref{fig:clm}A light blue class, good CLM) and one made of several distant clusters (\autoref{fig:clm}B light blue class, bad CLM)}. Moreover, classifiers require expensive training time \ma{(Appendix F)}.

%Another, less common way to evaluate clusterings is to examine how well the resulting clusters can be classified by classifiers \cite{abul03smc} or their ensemble \cite{rodriques18kbs}. 
%As classification scores can be compared across datasets, they could be used to validate CLM. However, as explained in the introduction, classifiers cannot detect certain cluster structures, are computationally expensive, and are not based on principled axioms related to clustering quality. 

A more direct approach is to average the results of multiple and diverse clustering techniques \cite{vega11jprai} as their high EVM scores would indicate a good CLM (\autoref{fig:clm}D). However, this approach is computationally expensive too \ma{(Appendix F)}. Moreover, the ground truth it \rev{approximates} is not based on principled axioms independent of any clustering technique, so it is likely biased in regard to the certain type of clusters these techniques can detect. For lack of a better option, though, we use this approach to \rev{get an approximate} ground truth in our experiments validating our axiom-based solution, while mitigating the bias by aggregating the EVM scores of multiple clustering techniques.

In contrast to classifiers or clustering techniques, most IVM are relatively inexpensive to compute \ma{(Appendix F)}. Also, as they are based on two criteria---\textit{compactness} (i.e., the pairwise closeness of data points within a cluster) and \textit{separability} (i.e., the degree to which clusters lie apart from one another) \cite{liu10idcm, tan05idm, zhao02cikm, rendon11ac}---they can examine the cluster structure in more details; in \autoref{fig:clm}, an IVM would give a higher score to partitions A and C than to B and F. 
Moreover, following the axiomatization of clustering by Kleinberg \cite{kleinberg02nips}, Ackerman and Ben-David \cite{bendavid08nips} proposed four within-dataset axioms that give a common ground to all IVMs: scale invariance, consistency, richness, and isomorphism invariance. These axioms set the requirements a function should satisfy to work properly as an IVM.

Nevertheless, IVMs were originally designed to compare and rank different partitions of the \emph{same} dataset as shown in \autoref{fig:clm}A-H. Therefore, IVM cannot be used to compare CLM \emph{across} different datasets in which not only the cluster structure but also the number of points, classes, and dimensions can vary (\autoref{fig:clm}I-L).
Here, we propose four additional axioms that an IVM should satisfy to allow this comparison, derive a new IVM satisfying them, and apply it to rank labeled datasets by their reliability to be used as a basis for clustering EVM.

\section{New Axioms for Internal Clustering Validation}

\label{sec:axiom}

%In this section, we first present Ackerman and Ben-David \cite{bendavid08nips}'s axioms that a function $f$ should satisfy to be a proper IVM. We call them \emph{within-dataset} axioms. Then, we introduce new between-dataset axioms.

\subsection{Ackerman and Ben-David's \emph{Within-dataset} Axioms}

Ackerman and Ben-David (\AB) proposed \textit{within-dataset} axioms \cite{bendavid08nips} that specify the requirements for IVM to properly evaluate clustering partitions. The first axiom is \textbf{W1: Scale Invariance}; it requires measures to be invariant to distance scaling. \textbf{W2: Consistency} is satisfied by a measure that increases when within-cluster compactness or between-cluster separability increases. \textbf{W3: Richness} requires measures to give any fixed cluster partition the best score over the domain by only modifying the distance function. Lastly, \textbf{W4: Isomorphism Invariance} ensures that an IVM does not depend on points identity. 
Detailed definitions are given in Appendix A.

\subsection{Axioms for Enabling \emph{Between-dataset} Comparison}

Within-dataset axioms do not consider the case of comparing scores across datasets; they assume the dataset is invariant. 
% and only the cluster partition changes. 
We propose four additional \emph{between-dataset} axioms that should be satisfied by internal validation measures to allow a fair comparison of cluster partitions across datasets.

\textbf{Notations }
% We use the same notation as \AB.
\rev{We follow the notations used in \AB.}
We define a finite domain set $X \subset \mathcal{D}$ of dimension $\Delta_X$, where $\mathcal{D}$ denotes data space.
We denote a clustering partition of $X$ as $C =\{C_1, C_2, \cdots, C_{|C|}\}$, where  $\forall i\neq j, C_i \cap C_j = \emptyset$ and $\cup_{i=1}^{|C|} C_i = X$. 
A distance function $d : \mathcal{D} \times \mathcal{D} \rightarrow \mathbb{R}$ is a function that satisfies $d(x, y) \geq 0$, $d(x, y) = d(y, x)$ and $d(x, y) = 0$ if $x=y$ for any $x, y \in \mathcal{D}$. 
%For two points $x, y \in X$, we say $x \sim_C y$ if $x$ and $y$ are in the same cluster and say $x \nsim_C y$ otherwise. 
If two point sets $X$ and $Y$ follow the same distribution, we say $X \overset{D}{=} Y$.
A measure is a function $f$ that takes $C, X, d$ as input and returns a real number. Throughout the section, higher $f$ implies better clustering. 
Additionally, we define $\underline{W}_\alpha$ a random subsample of the set $W$ ($\underline{W}_\alpha\overset{D}{=}W$) such that $|\underline{W}_\alpha|/|W|=\alpha$, and $\underline{C}_\alpha=\{\underline{C_i}_\alpha\}_{i\in 1\dots |C|}$.

\textbf{Goals and factors at play}
IVM$_{wthn}$ operate on fixed dataset $X$  with possible variations of $C$ and distance $d$. Hence, the number $|C|$ and sizes $|C_i|{\forall i}$ of the generated clusters can vary, while $X$ determines a common basis for comparison. \AB's \rev{within-dataset} axioms essentially state that the measure $f$ should be invariant to various aspects of the distance $d$. Hence, as $X$ is fixed, $f$ can only vary in relation to the clustering partition $C$. The satisfaction of the \AB's axioms is a way to ensure  \IVMw focus on measuring the clustering quality and nothing else.  
%Ackerman and Ben-David \cite{bendavid08nips} axioms aim to link \IVMw and their distance $d$ to some characteristics of natural clusters. nvariant to global scaling (W1) or to increase its value when density increase around its modes (W2). They also discuss the option for \IVMw to be invariant to the number of cluster labels $|C|$. As a consequence, \IVMw  quantify the matching between natural clusters (fixed) and generated clusters only depending on the label assignments $p(C_i|x)$. % (which determines labels' proportions $p(C_i)$).  

In contrast, between-dataset IVM (\IVMb) shall operate on varying $C$, $d$, and $X$. Imposing \IVMb to satisfy \AB's \rev{within-dataset} axioms will reduce the influence of $d$. Still  several aspects of the varying datasets $X$ now come into play and their influence on \IVMb shall be minimized. The sample size $|X|$ is one of them (\textbf{Axiom B1}) and the dimension $\Delta_X$ of the data another one (\textbf{Axiom B2}). 
Moreover, what matters is the \textit{matching} between natural clusters and data labels more than the number of clusters or labels; therefore, 
%Moreover, when considering labeled datasets where cluster labels are given rather than computed with a clustering technique,  
we shall reduce the influence of the number of labels $|C|$ (\textbf{Axiom B3}) imposed by the dataset. Lastly, we need to align \IVMb to a comparable range of values (\textbf{Axiom B4}) across datasets, in essence capturing all remaining hard-to-control factors unrelated to clustering quality (i.e., the \textit{matching} between natural clusters and data labels (CLM)) but integrated by the measure.

\textbf{Axiom B1} Invariance to the sample size is ensured if subsampling all clusters in the same proportion does not affect the \IVMb score, leading to the first axiom:

\textbf{\textit{Data-cardinality Invariance}} \textit{A measure $f$ satisfies data-cardinality invariance if
$\forall X,\forall d$ and for every clustering $C$ of $(X,d)$, $f(C,X,d) = f(\underline{C}_\alpha,X_\alpha,d)$ with $X_\alpha=\cup_{i=1}^{|C|} \underline{C_i}_\alpha$.}

% \hjc{A claim about "same distribution" is needed} \mac{No need, it is the case by definition of $X_\alpha$}

%$\forall X,\forall d$, and $\forall C, \forall C'$ partitions of $X$, $f(C,X,d)\leq f(C',X,d)  \Rightarrow f(\underline{C}_\alpha,\underline{X}_\alpha,d)\leq f(\underline{C'}_\alpha,\underline{X}_\alpha,d)$}

%\textit{A measure $f$ satisfies data-cardinality invariance if $\forall X, \forall d, \forall C$ over $(X,d)$, $f(C, X, d) = f(C', X', d)$ \mac{use a statement like if f1>f2 then f'1> f'2 or better use $\approx$ sign and do not claim theoretical proof but rather ideal principle that ourmeasures and tricks will realize in experiments}(where $X'$ is a randomly sampled subset of $X$, satisfying $X' \overset{d}{=} X$, and $C'$ is a clustering over $X'$ satisfying $\forall x, y \in C', x \sim_{C'}y$ for $ x \sim_C y$ and $x \nsim_{C'}y $ for $x \nsim_C y$, $|C_i| / |X| = |C'_i| / |X'|$, and $C_i \overset{D}{=} C'_i$ $\forall i \in \{1, 2, \cdots, |C|\}$)}.

\textbf{Axiom B2} We shall take into account that data dimension $\Delta_X$ may vary across datasets. An important aspect of the dimension called the concentration of distance phenomenon, which is related to the curse of the dimensionality \cite{bellman66science}, affects the distance measures involved in \IVMb. As dimension grows, the variance of data pairwise distances for any distribution tends to be constant while their mean value increases \cite{beyer99icdt, francois07tkde, lee11iccs}. Therefore, in high dimensional spaces,  $d$ will act as a constant function for any data $X$, thus an \IVMb $f$ will generate similar scores for all datasets. 
To mitigate this phenomenon, and as a way to reduce the influence of the dimension, we require the measures to be shift invariant \cite{lee11iccs, lee14cidm} so that the shift of the distances (i.e., growth of the mean) is canceled out:

\textbf{\textit{Shift Invariance }} A measure $f$ satisfies shift invariance if $\forall X, \forall d$ and for every clustering $C$ over $(X, d)$, $f(C, X, d) = f(C, X, d+ \beta)$ $ \forall \beta > 0$ where $d + \beta$ is a distance function satisfying $(d + \beta)(x,y) = d(x,y) + \beta$, $\forall x, y \in X$.

\textbf{Axiom B3} The number of classes should not affect an \IVMb, for example, two well clustered classes should get an \IVMb score as good as 10 well clustered classes. \AB proposed to aggregate class-pairwise \IVMw  to form other valid \IVMw. We follow this principle but state it as an axiom for \IVMb:

\textbf{\textit{Class-cardinality Invariance} } 
\textit{A measure $f$ satisfies class-cardinality invariance if $\forall X,  \forall d$ and $\forall C$ over $(X, d)$, $f(C, X, d) = $ }$ \textrm{agg}_{S \subseteq C, |S| = 2} f'(S, X, d)$ \textit{with}  $\textrm{agg}_S\in\{\textrm{avg}_S,\min_S,\max_S\}$ \textit{and $f'$ is an IVM.
}

By design, $f$ will satisfy all within or between axioms that $f'$ satisfies (Appendix B).

%\textit{A measure $f$ satisfies class-cardinality invariance if $\forall X, X',  \forall d, \forall C$ over $(X, d)$ and $\forall C'$ over $(X', d)$ satisfying $\sum_{S \subseteq C, |S| = 2} f(S, X, d) / {|C| \choose 2} = \sum_{S' \subseteq C', |S'| = 2} f(S', X', d) / {|C'| \choose 2}$, $f(C, X, d) = f(C', X', d)$. 

%\mac{is X' a subset of X? and C' a subset of C? not clear}\mac{JUSTIFY IT BETTER: intuition is if we have a measure which is invariant for a pair of classes, then we can use it for more than 2 classes by averaging over all pairs of classes, and it will generate a class-cardinality invariant measure.  We make sure the measure f is class-pairwise, then we divide the aggregate easure by the numebr of pairs. This is similar to computing a measure per class and divide by the number of classes. But the latter is not relevant because we need at least some class-pairwise comparison. So the xiom could be f must be an aggregate function of class-pairwise components, each of them being itself a valid between-dataset IVM. f'= $f'= aggregate_{i,j \in C}(f_{i,j}$} check section 5.4 of Bendavid paper to justify this axiom. For us there is no choice, we have to consider different number of classes because across datasets}

\textbf{Axiom B4} Lastly, we need to ensure that \IVMb  take a common range of values across datasets, so that their minimum and maximum values correspond to datasets with the worst and the best CLM, respectively, and that these extrema are aligned across datasets \ma{(we set them arbitrarily to 0 and 1)}:

\textbf{\textit{Range Invariance.}} \textit{A measure $f$ satisfies range invariance if $\forall X, \forall d$, and $C$ any clustering over $(X,d)$, $\min_{C} f(C,X,d)=0$ and $\max_{C} f(C,X,d)=1$.
}

\section{Generating Between-dataset Internal Validation Measures}

\label{sec:ivmbtw}

We propose technical tricks to generate \IVMb{}   \ma{that satisfy our supplementary axioms, and use these tricks} to generalize the Calinski-Harabasz index ($CH$) \cite{calinski74cis} to the between-dataset $CH$ index ($CH_{btwn}$).

%Here, we introduce how we designed IVM$_{btwn}$. We first present a generalization processes where each makes IVM satisfy a corresponding between-dataset axiom. Then, we show how Calinski-Harabasz index ($CH$) \cite{calinski74cis} is generalized to between-dataset Calinski-Harabasz ($CH_{btwn}$) through the processes. 

\subsection{Generalization Tricks \rev{for Enabling Between-dataset Comparison}}

% \mac{LEt's be more focused for each tricj independently of the others, one or two possible tricks per axiom without discussing much about how they relate to existing internal measures. Then in next section we apply these trick directly considering CH}

%Followings are the generalization processes that corresponds to each between-dataset axiom.

\textbf{Trick T1: Approaching data-cardinality invariance (B1) } We cannot guarantee the invariance of a measure for any subsampling of the data (e.g., very small sample size), but we can get some robustness to random subsampling if we use consistent estimators of population statistics  as building blocks of the measure, such as the mean or the median of a class, a pair of classes, or of the whole dataset, or quantities derived from them such as the average distance between all points of two classes. %. stable through sampling like the media, the average Typical statistics invariant through sampling are the average Point-cardinality invariance is fulfilled if the measure consists of point-cardinality invariant terms (i.e., terms having consistent value unless the overall distribution of the data points change), such as average within-cluster distance or distances between cluster centroids. Therefore, to achieve point-number invariance, we need to simply examine all terms in the measure and carefully substitute or remove the terms that are not point-cardinality invariant.

\textbf{Trick T2: Achieving shift invariance (B2) }
Considering a vector of distances $u=(u_1\dots u_n)$, we can define a shift invariant measure by using a ratio of exponential functions $g_j(u)=\frac{e^{u_j}}{\sum_k e^{u_k}}$. We observe that $\forall S\in \mathbb{R}, g_j(u+S) = \frac{e^{u_j+S}}{\sum_k e^{u_k+S}}=\frac{e^{u_j}}{\sum_k e^{u_k}}\frac{e^{S}}{e^{S}}=g_j(u)$, hence $g_j$ is shift invariant. This trick is at the core of the $t$-SNE loss function \cite{van08jmlr,lee11iccs}.
However, $g_j$ is not scale-invariant: $\forall \lambda\in \mathbb{R},g_j(\lambda u) = \frac{e^{\lambda u_j}}{\sum_k e^{\lambda u_k}}\neq\lambda g_j(u)$, hence it will not satisfy axiom W1. We can get back scale-invariance by normalizing each distance $u_i$ by a term that scales with all of them together, for example, their standard deviation: $\sigma(u)$. Now $g_j(\lambda u/\sigma(\lambda u))=g_j(\lambda u/\lambda \sigma(u))=g_j(u/\sigma(u))$ is both shift and scale invariant.

\textbf{Trick T3: Achieving class-cardinality invariance (B3) }
Class-cardinality invariance can be achieved by following the definition of Axiom B3, such as by defining the measure $f$ as the average of class-pairwise measures. %Formally, for an IVM $f$, we can make $f_c$ satisfying class-number invariance as
%$f_c(C, X, d) = \sum_{S \subseteq C, |S| = 2} f(S, X, d)$.
%Then, as $f_c(C, X, d) = f(C, X, d)$ if $|C| = 2$, 
%$f_c(C, X, d) = \sum_{S \subseteq C, |S| = 2} f_c(S, X, d)$
%and thus $f_c$ satisfies class-cardinality invariance.

%Ben-David and Ackerman \cite{bendavid08nips} showed that if $f$ satisfies within-dataset axioms, then $f_c$ also satisfies those axioms. We found that it is also true for between-dataset axioms (Appendix XX). Therefore, to check whether $f_c$ fulfills the remaining axioms, all we need to do it to prove that $f$ satisfies the axioms (Appendix YY).  

 %ratio defines the similarity between two points $x_i$ and $x_j$ as 
%$p_{i,j} = (e^{-d^2_{i, j}/ 2\sigma^2}) / (\Sigma_{k \neq l} e^{-d^2_{k, l} / 2\sigma^2})$,

\textbf{Trick T4: Achieving range invariance (B4) }
%We exploit normalization to make measures not to be biased by the distribution. In detail, normalization of a measure $f$ is defined as $(f - E(f)) / (\max(f) - E(f))$ \cite{wu09kdd}; the process enables fair comparison between datasets with different data distribution by aligning max/min score between datasets. 
A common approach to get a unit range for $f$ is to use min-max scaling $f_u=(f-f_{\min})/(f_{\max}-f_{\min})$.
However, determining the possible minimum and maximum values of $f$ for any data $X$ is not straightforward. Theoretical extrema are usually computed for edge cases far from realistic $X$ and $C$. Wu et al. \cite{wu09kdd}  propose to estimate the worst score over a given dataset $X$ by the expectation $\hat{f}_{\min}=E_{\pi}(f(C^{\pi},X,d))$ of $f$ computed over random partitions $C^{\pi}$ of $(X,d)$ preserving class proportions $|C^{\pi}_i|=|C_i|{\forall i}$ (Trick 4a)---arguably the worst possible clustering partitions of $X$. In contrast, it is hard to estimate the maximum achievable score over $X$, as this is the very objective of clustering techniques. If the theoretical maximum $f_{\max}$ is known and finite, we propose to use it by default; otherwise, if infinite, we propose to use a logistic function (Trick 4b) to scale it down to $1$ (Note that the scaled measure $f_u$ is $0$ if $f_{\max}\rightarrow+\infty$).

\subsection{Generalizing the Calinski-Harabasz Index}

\label{sec:gener}

As a proof-of-concept, we use the proposed tricks to generalize the $CH$ index to the $CH_{btwn}$ index that satisfies \ma{both within-dataset (W) and between-dataset (B)} axioms. We select $CH$ as it is a representative  \IVMw \cite{liu10idcm, liu13tc,  maulik02tpami, xiong13cvm} widely used for clustering evaluation \cite{wang19mse, lukasik15cec, baarsch12imecs}. It is defined as:
\[
 CH(C, X, d) = \displaystyle\frac{\sum_{i=1}^{|C|}|C_i|d^2(c_i, c) / (|C| - 1)}{\sum_{i=1}^{|C|} \sum_{x \in C_i} d^2(x, c_i) / (|X| - |C|)}, 
\]
where $c_i$ is the centroid of $C_i$ and $c$ is the \rev{centroid} of $X$. A higher value implies a better CLM. The denominator and numerator measure compactness and separability, respectively. Both are estimators of population statistics (Trick 1), reducing by design the influence of data-cardinality (Axiom B1). %, As $CH$ is already in fractional form, there is no need to apply the process for point-cardinality invariance. 
We get shift invariance (Axiom B2) while preserving scale invariance (Axiom W1) by substituting the square distances by their exponential form normalized by the standard deviation $\sigma_{d}$ of the distances of data points to the centroid (Trick 2), leading to: 
\[
 CH_1(C, X, d) = \frac{\sum_{i=1}^{|C|}|C_i|e^{d(c_i, c) / \sigma_{d}} / (|C| - 1)}{\sum_{i=1}^{|C|} \sum_{x \in C_i} e^{d(x, c_i) / \sigma_{d}} / (|X| - |C|)}.
\]

Then, we apply min-max scaling (Axiom B4). 
As $\max(CH_1) = +\infty$, we transform it through a logistic function (Trick 4b) $CH_2 = 1 / (1 + CH_1^{-1})$ so $CH_{2\max}=1$. We estimate the worst score as the expectation of $CH_2$ over random clustering partitions $C^\pi$ (Trick 4a): $CH_{2\min}=E_\pi(CH_2(C^{\pi}, X, d))$. We get $CH_3= (CH_2 - CH_{2\min}) / (CH_{2\max} - CH_{2\min})$.  %Then, as $CH_2$ ranges from 0 to 1, the normalized measure can be defined as $CH_3 = (CH_2 - E(CH_2)) / (\max(CH_2) - E(f))$.
%Note that $E(CH_2)$ is computed by repeating the measurement with random shuffled class labels and taking the average. Formally, $E(CH_2) = E[CH_2(C^{\pi}, X, d)]$ with $C^{\pi}$  a random permutation of $C$, where $|C^{\pi}_i|=|C_i| \forall i$ to preserve class proportions. 

Lastly, we satisfy class-cardinality (Axiom B3) by averaging class-pairwise scores (Trick 3), which determines the between-cluster Calinski-Harabasz index:
\[
CH_{btwn}(C, X, d) = \frac{1}{{|C| \choose 2}}\sum_{S \subseteq C, |S| = 2} CH_3(S, X, d). 
\]

Unlike $CH$, which misses all between-dataset axioms except B1, $CH_{btwn}$ satisfies all of them by design, and we prove it also satisfies all within-dataset axioms (Appendix B).

The existence of at least one \IVMb provides evidence pointing toward the consistency of our axioms. Still, we cannot prove their completeness nor their soundness for lack of a clear definition of what a good clustering is (See \AB \cite{bendavid08nips} for a discussion of these concepts for clustering).
Our experiments validate the importance of these axioms for comparing the CLM of different datasets.

In terms of computational complexity, 
\rev{$CH$, $CH_1$, and} $CH_2$ are $O(|X|\Delta_X)$, thus $CH_{2\min}=O(|X|\Delta_X T)$ while $CH_{2\max}=O(1)$, where $T$ is the number of Monte Carlo simulations to estimate $CH_{2\min}$. Hence, $CH_3=O(|X|\Delta_X T)$, and finally  $CH_{btwn}=O(|X|\Delta_X T P_C)$, where $P_C=|C|(|C|+1)/2$ is the number of pairs of classes.
Worst-case complexity of $CH_{btwn}$ is linear with all parameters but quadratic with the number of classes, making it very scalable (Appendix F).

\section{Evaluation}

%We conducted a series of quantitative experiments to evaluate $CH_{btw}$. 
%While validity analysis experimentally verified that $CH_{btw}$ satisfies the between-dataset axioms, rank correlation analyses confirmed $CH_{btw}$'s performance in evaluating and comparing CLM, both in between- and within-dataset. 

\subsection{Ablation Study of Between-dataset Calinski-Harabasz index}
\label{sec:validity}

\begin{figure}[t]
    \centering
    \includegraphics[width=\textwidth]{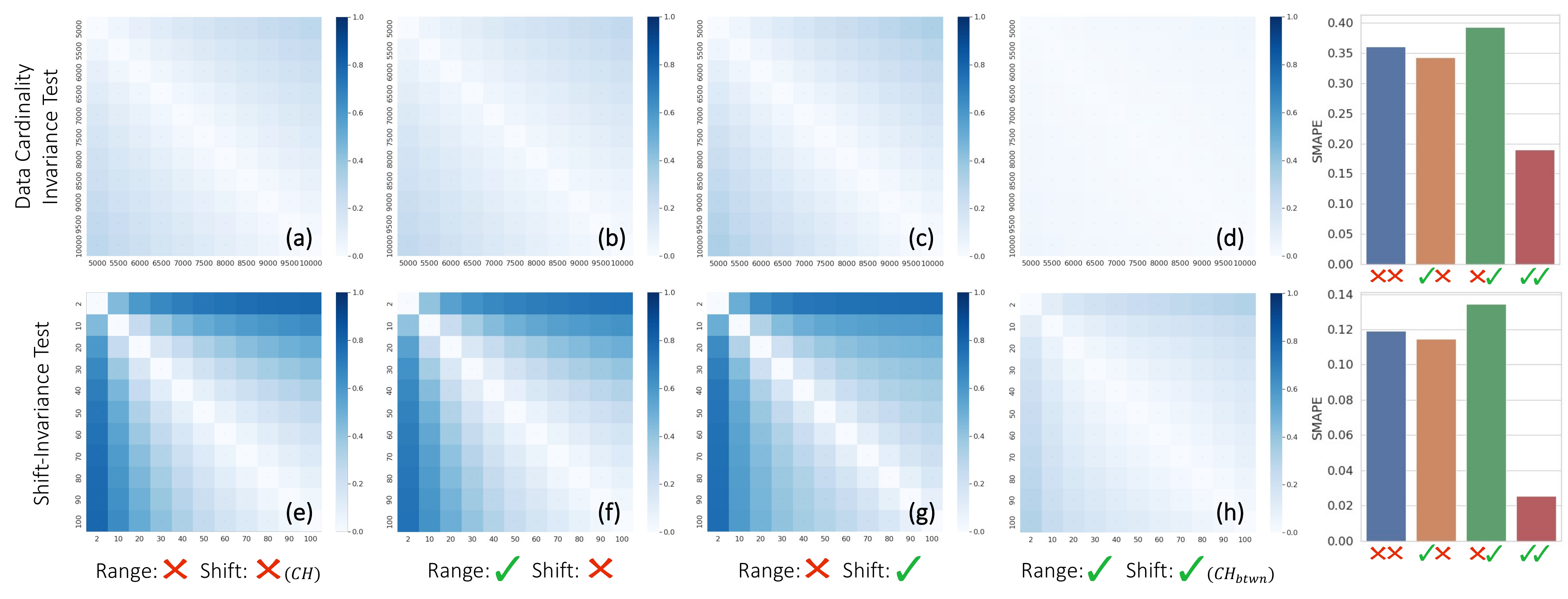}
    \vspace{-4mm}
    \caption{
    Ablation study of \CHb (\autoref{sec:validity}). Heatmaps (a--h) show the SMAPE of $CH_v$ variants (each column) for all pairs $(N_a,N_b)$ of controlled dataset sizes (top row) and pairs $(\Delta_a,\Delta_b)$ of controlled dimensions (bottom row). The lighter the color, the lower the error and the less sensitive $CH_v$ to variations of the controlled factor. Bar charts (right) show the average over all pairs of values of controlled factors for each $CH_v$ variant. \rev{See Appendix H for high-resolution images.} }
    \label{fig:val_results}
    \vspace{-2mm}
\end{figure}

\textbf{Objectives and design}
\CHb derives from $CH$ by using tricks T2, T3, and T4. We want to evaluate the role \ma{these tricks play in making \IVMb satisfy the new axioms. We consider a synthetic dataset from a previous study \cite{abbas19cgf, AupetitVIS19} made of two bivariate Gaussian clusters (class labels) with various levels of CLM, to which we add noisy dimensions.} We consider four variants \CHv of \CHb with shift (T2) and range (T4) tricks switched \textit{on} or \textit{off} ($CH_v\in\{CH,CH_{T2},CH_{T4},CH_{T2\&T4}=CH_{btwn}\}$). \rev{The effect of T1 is not evaluated as B1 is already satisfied by both $CH$ and $CH_{btwn}$, and the effect of} T3 is not evaluated because \rev{the ground truth synthetic datasets contain only two classes}. We control the cardinality (B1) and dimension (B2) of the datasets to evaluate how sensitive \CHv variants are to \ma{variations of these conditions (the lower, the better)}. \rev{We do not control class-cardinality (B3) as the number of classes (2) is imposed by the available data. Range-invariance (B4) is not controlled as it is imposed by the min-max trick (T4) and not a characteristic of the datasets.}

 %two within-dataset axioms (scale invariance, consistency) and three between-dataset axioms (cardinpoint-number invariance, shift invariance, baseline invariance); these axioms are tested as they define how the measures should behave. Richness, isomorphism invariance, and cluster-number invariance are close to theoretical requirements and thus not testable.

%\subsubsection{Design}

\textbf{Datasets} We prepared 1,000 base datasets $\{X_1\dots X_{1000}\}$, each one consisting of $|X|=10,000$ points sampled from two Gaussian clusters ($|C|=2$) within the 2D space and augmented with $98$ noisy dimensions ($\Delta=100$). We controlled the eight independent parameters (ip) of the Gaussians: two covariance matrices (3 ip each), class proportions (1 ip), and the distance between Gaussian means (1 ip), following a previous study \cite{abbas19cgf, AupetitVIS19} (see figure in Appendix C). We add Gaussian noise along the supplementary dimensions, to each cluster-generated data, with a mean 0 and a variance equal to the minimum span of that cluster's covariance. We generated any dataset $X_{i,t}$ by specifying a triplet $(X_i,N_t,\Delta_t)$ with $X_i$ a base dataset, $N_t$ the number of data randomly sampled from $X_i$ preserving cluster proportions, and $\Delta_t$ its dimension where the first two dimensions always correspond to the 2D cluster space.
\textbf{Sensitivity to data-cardinality (B1)} (\autoref{fig:val_results} top) 
For each of the $1000$ base data $X_i$, we generated $11$ datasets $X_{i,t}=(X_i,N_t,\Delta_t)_{i\in[1\dots1000],t\in[0\dots10]}$ with the controlled data cardinality set to $N_t=500t+5000$ and $\Delta_t$ drawn uniformly at random from $[2,\dots ,100]$.
\textbf{Sensitivity to dimensionality (B2)} (\autoref{fig:val_results} bottom)
For each of the $1000$ base data $X_i$, we generated $11$ datasets $X_{i,t}=(X_i,N_t,\Delta_t)_{i\in[1\dots1000],t\in[0\dots10]}$ with $N_t$ drawn uniformly at random from $[500,\dots,5000]$ and the controlled dimension set to $\Delta_0=2$ or $\Delta_t=10t, \forall t>0$.

\textbf{Measurements} 
% We use the Euclidean distance for $d$ and the class labels $C$ given by the two Gaussian clusters of each base dataset. 
For each $CH_v$, we compute the matching between a pair ($a,b)$ of values of the controlled factor $t$ (\textit{e.g.} ($\Delta_a,\Delta_b)=(10,30)$) across all $1000$ base data using: $S_{k\in[1\dots 1000]}(CH_v(C,X_{k,a},d),CH_v(C,X_{k,b},d))$, where S is the Symmetric Mean Absolute Percentage Error (SMAPE) \cite{tofallis15jors} adapted to compare measures with different ranges: ${\displaystyle {S_{k\in K} (F_k,G_k)}={\frac {1}{n}}\sum _{k\in K}\frac{|F_{k}-G_{k}|}{|F_{k}|+|G_{k}|}}$ (0 best, 1 worst).

\textbf{Results} 
\autoref{fig:val_results} shows that all $CH$ variants are slightly sensitive to changes in the data size (a--d), with a larger difference of size leading to bigger errors (off-diagonal darker shades of blue). The average error is about $10\%$ for all variants except \CHb (top row, blue, orange, and green bars), and \CHb is five times less sensitive to data cardinality than any other variant (top row, red bar). Regarding dimensionality (e--h), all variants except \CHb (h) are more strongly affected by larger differences in dimension, with about $35\%$ error on average, while \CHb (red bar) is slightly below $20\%$ on average, a two-fold improvement over other variants. 

The bar chart shows that the combination of both shift invariance (T2) and range invariance (T4) tricks is necessary to get \CHb satisfying axioms B1 (cardinality invariance) and B2 (shift invariance). It is unexpected, though, that using the shift invariance trick alone makes $CH_v$ more sensitive to the dimension. However, this can be explained by the fact that the exponential trick cancels the global shift of all distances (what it is designed for), disregarding the effect on the range of the IVM itself (a non-linear aggregation of distances), a factor that is then mitigated by the range trick (T4).

\subsection{Between-dataset Rank Correlation Analysis}

\begin{table}[t]
    \centering
    \scalebox{0.77}{
    \begin{tabular}{crrrrr}
    \toprule 
    %& & \multicolumn{3}{c}{} \\ 
   % \cmidrule(r){3-5}
    \ma{Spearman's rank correlation}& &\multicolumn{4}{c}{\ma{GT-ranking EVMs}}  \\     
    \ma{with approximate ground truth CLM}& & \texttt{ami} & \texttt{arand} & \texttt{vm} & \rev{\texttt{nmi}} \\ 
    \midrule
    \multirowcell{8}{Classifiers } & SVM & 0.5427 & 0.6235 & 0.4625 & 0.4827\\   
                                 & $k$NN & 0.4876 & 0.5810 & 0.3974 & 0.4094\\ 
                                 & MLP & 0.4405 & 0.5386 & 0.3600  & 0.3761\\ 
                                 & NB & 0.4126 & 0.5276 & 0.3157  & 0.3130\\ 
                                 & RF & 0.4893 & 0.5741 & 0.3991 & 0.3889\\ 
                                 & LR & 0.4456 & 0.5382 & 0.3666  & 0.3873\\ 
                                 & LDA & 0.4999 &  0.5726 & 0.3945 & 0.3606\\ 
                                 & Ensemble & 0.5922 & 0.6748 & 0.4614  & 0.4099\\ 
    \midrule
         \multirowcell{6}{\IVMw}  
          & \rev{Silhouette} & 0.5648 & 0.6800 & 0.4549 &   0.4208\\
          & \rev{Xie-Beni}   & 0.6201 & $^{*}$0.7019 & 0.4934 & 0.4446\\
          & Dunn             & 0.4026    & 0.3534 & 0.5366 & $^{*}$0.5979\\
          & I Index          & 0.5668  & 0.5957 & $^{**}$0.6086 &  $^{**}$0.6454\\
          & Davies-Bouldin   & $^{**}$0.7091 & $^{**}$0.7513 &  $^{*}$0.5719 & 0.5015\\ 
          & $CH$             & $^{*}$0.5923& 0.6222 & 0.4487 & 0.3810\\
    \midrule
           \IVMb (ours)  & $CH_{btwn}$   & $^{***}$0.7893& $^{***}$0.7981 & $^{***}$0.7022  & $^{***}$0.6561 \\
 
    \bottomrule
    \addlinespace[0.115cm]
    \multicolumn{5}{l}{
        \footnotesize
        $^{***}$, $^{**}$, $^{*}$: first, second, and third highest scores \ma{for each EVM}  
    } \\
    \multicolumn{5}{l}{
        \footnotesize
        Every result was validated to be statistically significant ($p < .001)$ \rev{through Spearman's rank correlation test.}
    }
    \end{tabular}
    }
    \vspace{2.5mm}
    \caption{
    Rank correlations between \rev{approximate} ground truth CLM ranking \ma{based on 9 clustering techniques} and estimated CLM ranking obtained by \CHb, various \IVMw, and classifiers. 
    $CH_{btwn}$ rankings (***) outperform \rev{all the competitors and achieved an improvement of about 20\% compared to its within-dataset version ($CH$).}  }
    \label{tab:dataset_entire}
    \vspace{-3mm}
\end{table}

\label{sec:bdsa}

\textbf{Objectives and design}
We  assess $CH_{btwn}$ against competitors for best estimating the CLM ranking of publicly available labeled datasets.
We \rev{approximate} a ground truth CLM quality for each labeled dataset using multiple clustering techniques. We then compare the rankings made by all competitors and $CH_{btwn}$ to this ground truth using Spearman's rank correlation. 
%Spearman's rank correlation.

\textbf{Datasets} We collected 96 publicly available labeled datasets with diverse numbers of data points\rev{,} class labels, cluster patterns (presumably), and dimensionality (Appendix E).

\textbf{\rev{Approximating the ground truth} CLM} \rev{For lack of definite ground truth clusters in multidimensional real data,} we used the maximum EVM score achievable by \ma{nine} various clustering techniques on a labeled dataset as \rev{an approximation of the} ground truth (GT) CLM score for that dataset. These GT scores were used to get the GT-ranking of all the datasets. 
This scheme relies on the fact that high EVM implies good CLM (\autoref{sec:intro}; \autoref{fig:clm} A and D).
We used Bayesian optimization \cite{snoek12neurips} to find the best hyperparameter setting for each clustering technique. 
We obtained GT-ranking based on \ma{four} EVMs: adjusted rand index (\texttt{arand}) \cite{santos09icann}, adjusted mutual information (\texttt{ami}) \cite{vinh10jmlr}, V-measure (\texttt{vm}) \cite{rosenberg07emnlp}, \rev{and normalized mutual information (\texttt{nmi}) \cite{strehl02jmlr} with geometric mean}. 
For clustering techniques, we used HDBSCAN \cite{campello13akddm}, DBSCAN \cite{schubert17tds},  $K$-Means \cite{hartigan79jstor}, $K$-Medoids  \cite{park09esa}, $X$-Means \cite{pelleg00icml}, Birch \cite{zhang96sigmod}, and Single, Average, and Complete variants of Agglomerative Clustering \cite{mullner11arxiv} (Appendix D).  

\textbf{Competitors} We compared supervised classifiers, \IVMw,  and  $CH_{btwn}$ to the GT ranking. For classifiers, we used SVM, $k$NN, Multilayer Perceptron (MLP), Naive Bayesian Networks (NB), Random Forest (RF), Logistic Regression (LR), Linear Discriminant Analysis (LDA), and their ensembles; \rev{the selected classifiers are the ones used for evaluating clustering in  Rodr\'iguez et al. }\cite{rodriques18kbs}. We measured the classification score of a given labeled dataset, using five-fold cross validation and Bayesian optimization \cite{snoek12neurips} to find the best hyperparameter setting. The accuracy in predicting class labels was averaged over the five validation sets to get a proxy of the CLM score for that dataset. For the ensemble method, we got the proxy as the highest accuracy score among the seven classifiers for each dataset independently. \rev{Regarding \IVMw, we considered the list of Liu et al. \cite{liu10idcm}, except the ones optimized based on the elbow rule (\textit{e.g.}, Modified Hubert $\Gamma$ statistic \cite{hubert85classification})
and the ones requiring several clustering results (\textit{e.g.}, S\_Dbw index \cite{halkidi01icdm}), thus we used:} \CH, Davies-Bouldin index \cite{davies79tpami}, Dunn index \cite{dunn74joc}, I index \cite{maulik02tpami}, Silhouette \cite{rousseeuw87jcam}, and Xie-Beni index \cite{xie91tpami} (See details in Appendix D).

%\subsubsection{Results and Discussions}
\textbf{Results} 
\autoref{tab:dataset_entire} shows that for every EVM, $CH_{btwn}$ (***) outperforms all \rev{competitors. Especially, $CH_{btwn}$ achieved a performance improvement of about 20\% compared to $CH$. The second (**) and third (*) places  vary depending on the EVM, but they are all part of the \IVMw category.}
% second-best competitors by about $10\%$ with up to a $0.8$ rank correlation with GT ranking in all cases. The EVM used to compute the GT does not affect this ranking. Davies-Bouldin and \CH are the best \IVMw with about $0.7$, while other \IVMw and, as expected, all classifiers are near $0.5$.  
Therefore, $CH_{btwn}$ can be used as a reliable measure \ma{of CLM to rank datasets} (\autoref{fig:btw_result}) despite their drastic variations in terms of dimension, number of class labels, and data size. It also runs far faster  than optimizing any of the GT clustering techniques (tens of seconds versus several hours for all 96 datasets; Appendix F), clearly demonstrating its benefit \ma{both in terms of time and accuracy}.

% \mac{OK, done. it would be good to give a factor of speed increase, like it is twice faster or ten times faster... How many runs of Monte-Carlo simulations are sufficient to get a good estimate of Fmin? Maybe it can be accelerated by setting directly the centroid of all classes to the one of the data? You used this trick in some formula I remember, maybe not easy to prove, but maybe a practical approach to use to shorten the computation time?}
% \hjc{I checked it right before, and because of expectation (for normalization), it is no more faster than clustering tehcniques like DBSCAN. But it is still faster than "optimizing clustering techniques". Maybe we should argue that "It also runs faster than optimizing any of the GT clustering techniques."}
% \hjc{Currently, 10,000 point version of section 5.1 (almost using every processors). test is running. I'll check the time after this test ends }

\section{Application: Ranking the Labeled Datasets for Reliable EVM}

\label{sec:selection}

\begin{figure}[t]
    \centering
    \includegraphics[width=\textwidth]{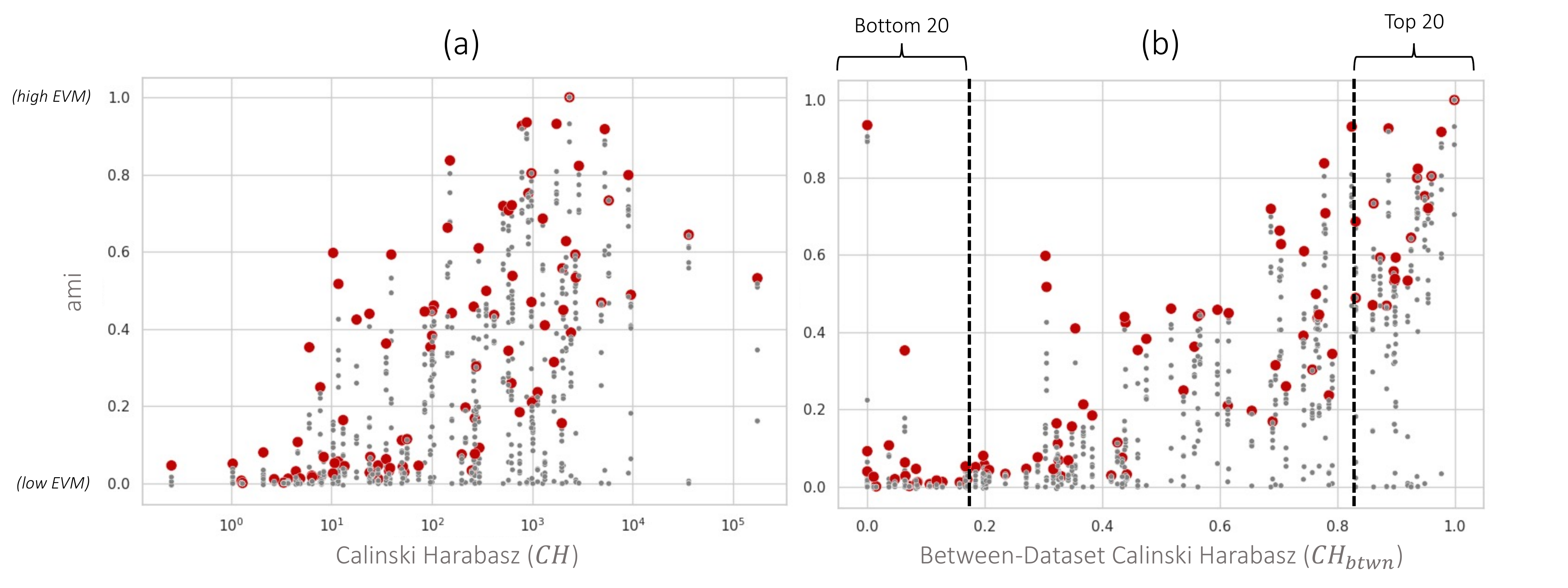}
    \vspace{-3mm}
    \caption{
    % \mac{dispay side by side CH and CHbtwn plots for visual comparison. Show bottom 20 and top 20. remove all ABCD and black dashed lines and remove all a,b,c,d. keep only one of the red dashed line to explain what is displayed}
    All (gray points) and best (red points) \texttt{ami} scores of GT clustering techniques for the $96$ benchmark datasets, ranked by \CH (left) and \CHb (right). \ma{The top $20$ datasets in terms of} $CH_{btwn}$ (right) are \ma{the most reliable to evaluate and compare clustering techniques using EVMs.} % \maa{The top left outlier (b) is a hard-to-cluster dataset with correct labels (\autoref{fig:clm}J), as it is detected by GT clustering techniques (high EVM, \autoref{fig:clm}D) but missed by our CLM proxy \CHb.}  
    % \mac{Highlight the max EVM points for each dataset, making it bigger, because this is what is used in Table 1 and what we shall focus on here. Can we display another scatterplot with only the max EVM points all in black color? It would help decide on a good threshold. Can we match the letters ABCD here to the ones used in Figure 1? We should refer to Figure 1 here to help interpretation. Highlight the edges of the corner B, align C horizontally with D both at the bottom, and align B vertically with D both on the right, so all letters are at the corner of the plot. Color the letters in blue so they stand out or make them bold font. You annotate datasets a,b,c,d but only d is mentioned in the caption. Refer to seciton in text where to read more explanations. Give clear title to the figure related to its aim: "Selecting the best benchmark datasets for EVM"}
    }
    \label{fig:btw_result}
    \vspace{-2mm}
\end{figure}

\begin{figure}
    \centering
    \includegraphics[width=\textwidth]{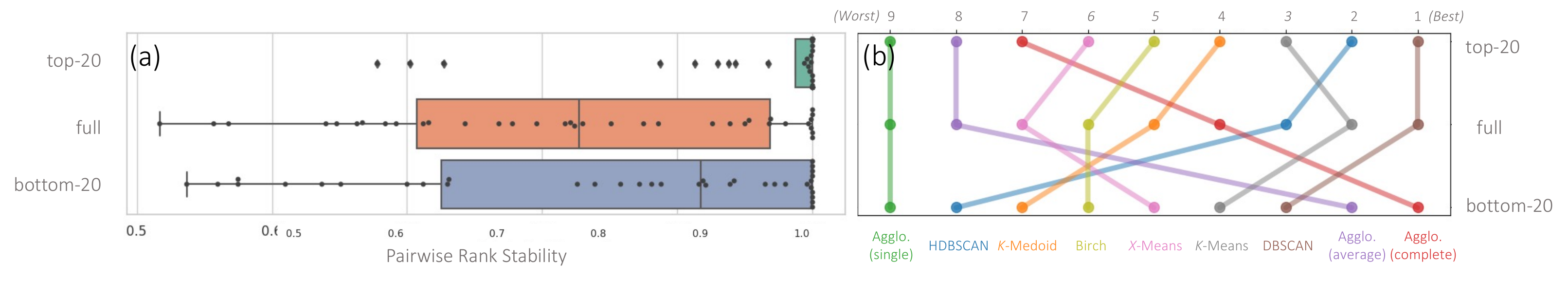}
        \vspace{-3mm}
    \caption{(a) 
    % \mac{Can we align the two plots on the y-axis so transpose the right plot? it would ease visual conection between them}
     Distribution of pairwise rank stability for \texttt{bottom-20} (blue; $\mathcal{P}^-$), \texttt{full} (orange; $\mathcal{P}^*$), and \texttt{top-20} (green; $\mathcal{P}^+$) datasets. (b) 
     Rankings of clustering techniques for each set. All rankings are based on \texttt{ami} EVM averaged over the datasets within each set. Using the datasets top-ranked by \CHb as a proxy of their good CLM leads to stable and reliable rankings ((a) green bar). } 
    \label{fig:rank_analysis}
    \vspace{-2mm}
\end{figure}

\textbf{Objectives and design} We want to demonstrate the importance of \maa{evaluating} the CLM of benchmark datasets prior to conducting the external validation of clustering techniques. 
Here, in addition to the \texttt{full} set of $96$ public 
datasets ($\mathcal{P}^*$), we consider the \texttt{top-20}  ($\mathcal{P}^+$) and \texttt{bottom-20}  ($\mathcal{P}^-$) datasets as per their \CHb rank (\autoref{fig:btw_result}b) (the \texttt{top-20} and \texttt{bottom-20} datasets are given in Appendix C).

We consider simulating the situation where a data scientist would arbitrarily choose $10$ benchmark datasets ($\mathcal{B}$) among the datasets at hand for the task T of ranking clustering techniques according to $EVM_{\mathcal{B}}$, the average EVM over $\mathcal{B}$.
For each $\mathcal{P}\in\{\mathcal{P}^+,\mathcal{P}^*,\mathcal{P}^-\}$, we simulate $100$ times picking $\mathcal{B}$ at random among $\mathcal{P}$. For each $\mathcal{P}$, we measure the pairwise rank stability $P_\mathcal{B}(A,B)=\max(1-p,p)$ of clustering techniques A and B over $\mathcal{B}$ by counting the proportion $p$ of cases $\texttt{ami}_{\mathcal{B}}(A)>\texttt{ami}_\mathcal{B}(B)$.

\textbf{Assumptions} We expect that conducting T on any subset of good-CLM datasets would provide similar rankings (\autoref{fig:clm}A) where pairwise ranks remain stable ($\forall (A,B), P_\mathcal{B}(A,B)\approx 1$), whereas conducting T using bad-CLM datasets would lead to arbitrary and unstable rankings ($\forall (A,B), P_\mathcal{B}(A,B)$ spread over $[0.5, 1]$) %due to the inconsistency between their clusters and unreliable-ground-truth labels
(\autoref{fig:clm}BEH). 

\textbf{Results and discussion} \autoref{fig:rank_analysis}a shows that pairwise ranks stay stable only in $\mathcal{P}^+$, which verifies our assumptions. 
Moreover, we found that the rankings of clustering techniques made by $EVM_{\mathcal{P^+}}$, $EVM_{\mathcal{P^*}}$, and $EVM_{\mathcal{P^-}}$ are completely different (\autoref{fig:rank_analysis}b). 
%Therefore, \maa{ignoring the CLM of datasets may} lead to erroneous conclusions when benchmarking clustering techniques.
\rev{Still, some datasets within $\mathcal{P}^-$ (\textit{e.g.}, Spambase, Hepatitis \cite{asuncion07uci}) 
have been used for external clustering validation in previous studies \cite{rehioui16pcs, khan21icecit, monath17nips, monath19kdd}}
\maa{without CLM evaluation, casting doubt on their conclusion and showing this issue shall gain more attention in the benchmarking community.} 
 \maa{CLM scores could be used further to inform benchmarking results (Appendix G) or to improve dataset's reliability by modifying datasets' class labels.}

\section{Conclusion}

In this research, we provided a grounded way to \maa{evaluate the reliability of} benchmark labeled datasets used for the external evaluation of clustering techniques. \maa{We proposed to measure their level of cluster-label matching (CLM). }
We presented four between-dataset axioms and  technical tricks to generate measures that satisfy them. We used these tricks to design a new between-dataset internal validation measure \CHb generalizing the Calinski-Harabasz index for across-datasets comparisons. We studied the accuracy of this measure to rank $96$ benchmark datasets and showed that it outperforms all competitors in terms of time and accuracy. We demonstrated its usefulness in determining the most reliable datasets for comparing clustering techniques.

As future work, we want to explore further the use of our tricks to generalize other \IVMw, \maa{ and explore how to use the CLM score to build better clustering benchmarks.}

% Due to their natural purpose, internal clustering measures are not able to be used to compare clusterings of different datasets. To tackle this limitation, we introduced clusterness measures, which generalizes internal measures to be used across arbitrary datasets, thus enabled several helpful use scenarios. Our experiments verified clusterness measures' applicability to one of these---to use clusterness score as an difficulty of dataset to be classified or clustered.

% As a future work, we would like to test other possible application scenarios for our measure. For example, using clusterness measure as an input feature of Machine Learning system regarding class labels or clustering will be an interesting direction to go.

% %%%%%% Comment out for the main submission
% \begin{ack}
% This work was supported by AAA, BBB, and CCC.
% \end{ack}

\bibliographystyle{unsrt}

{
\small
\bibliography{ref}
}

\end{document}